\documentclass[journal,twoside,web]{ieeecolor}
\usepackage{cite}
\usepackage{amsmath,amssymb,amsfonts}
\usepackage{algorithmic}
\usepackage{graphicx}
\usepackage{textcomp}
\usepackage{adjustbox}

\usepackage{float}
\usepackage{caption}
\usepackage{subcaption}
\usepackage{multirow}
\usepackage{mathtools}
\usepackage{commath}
\usepackage{booktabs}
\usepackage{bbm}
\usepackage[linesnumbered,ruled, vlined]{algorithm2e}

\def\logoname{LOGO-tmi-web}
\definecolor{subsectioncolor}{rgb}{0,0,0}
\def\BibTeX{{\rm B\kern-.05em{\sc i\kern-.025em b}\kern-.08em
    T\kern-.1667em\lower.7ex\hbox{E}\kern-.125emX}}
\begin{document}
\title{Artifact-Tolerant Clustering-Guided Contrastive Embedding Learning for Ophthalmic Images}

\author{Min Shi, Anagha Lokhande, Mojtaba S. Fazli, Vishal Sharma, Yu Tian, Yan Luo, Louis R. Pasquale, Tobias Elze, Michael V. Boland, Nazlee Zebardast, David S. Friedman, Lucy Q. Shen, Mengyu Wang
\thanks{Min Shi, Anagha Lokhande, Mojtaba S. Fazli, Vishal Sharma, Yu Tian, Yan luo, Tobias Elze and Mengyu Wang (corresponding author) are with the Harvard Ophthalmology AI Lab, Schepens Eye Research Institute of Massachusetts Eye and Ear, Harvard Medical School, Boston, MA, USA (e-mails: \{mshi6, alokhande, mfazli, vsharma12, ytian11, yluo16, tobias\_elze, mengyu\_wang\}@meei.harvard.edu).}
\thanks{Louis R. Pasquale is with the Eye and Vision Research Institute, Icahn School of Medicine at Mount Sinai, New York, NY, USA (e-mail: louis.pasquale@mssm.edu).}
\thanks{Michael V. Boland, Nazlee Zebardast, David S. Friedman and Lucy Q. Shen are with the Massachusetts Eye and Ear, Harvard Medical School, Boston, MA, USA (e-mails: \{michael\_boland, nazlee\_zebardast, david\_friedman, lucy\_shen\}@meei.harvard.edu).}
}

\maketitle

\newcommand{\modelname}{EyeLearn}
\newcommand{\embeddim}{d}
\newcommand{\ncluster}{C}
\newcommand{\data}{\mathbf{D}}
\newcommand{\datax}{\mathbf{X}}
\newcommand{\datay}{\mathbf{Y}}
\newcommand{\datasize}{|\mathbf{D}|}
\newcommand{\spaceR}{\mathbf{R}}
\newcommand{\embedspace}{\mathcal{H}}
\newcommand{\embed}{\mathbf{h}}
\newcommand{\weightw}{\mathbf{W}}
\newcommand{\membank}{\mathbf{B}}
\newcommand{\bankfeat}{\mathbf{V}}
\newcommand{\bankclust}{\mathbf{C}}
\newcommand{\width}{W}
\newcommand{\lab}{c}
\newcommand{\height}{H}
\newcommand{\banksize}{M}
\newcommand{\masks}{\mathbf{Z}}
\newcommand{\maskm}{\mathbf{M}}
\newcommand{\identity}{\mathbf{I}}
\newcommand{\scaler}{r}
\newcommand{\biasb}{b}
\newcommand{\layerl}{l}
\newcommand{\loss}{\mathcal{L}}
\newcommand{\batchsize}{K}
\newcommand{\dist}{\mathbf{u}}
\newcommand{\batchembeds}{\mathbf{H}}
\newcommand{\numnegs}{N}

\begin{abstract}
Ophthalmic images and derivatives such as the retinal nerve fiber layer (RNFL) thickness map are crucial for detecting and monitoring ophthalmic diseases (e.g., glaucoma). For computer-aided diagnosis of eye diseases, the key technique is to automatically extract meaningful features from ophthalmic images that can reveal the biomarkers (e.g., RNFL thinning patterns) linked to functional vision loss. However, representation learning from ophthalmic images that links structural retinal damage with human vision loss is non-trivial mostly due to large anatomical variations between patients. The task becomes even more challenging in the presence of image artifacts, which are common due to issues with image acquisition and automated segmentation. In this paper, we propose an artifact-tolerant unsupervised learning framework termed \text{\modelname} for learning representations of ophthalmic images. \text{\modelname} has an artifact correction module to learn representations that can best predict artifact-free ophthalmic images. In addition, \text{\modelname} adopts a clustering-guided contrastive learning strategy to explicitly capture the intra- and inter-image affinities. During training, images are dynamically organized in clusters to form contrastive samples in which images in the same or different clusters are encouraged to learn similar or dissimilar representations, respectively. To evaluate \text{\modelname}, we use the learned representations for visual field prediction and glaucoma detection using a real-world ophthalmic image dataset of glaucoma patients. Extensive experiments and comparisons with state-of-the-art methods verified the effectiveness of \text{\modelname} for learning optimal feature representations from ophthalmic images.
\end{abstract}


\section{Introduction}
\IEEEPARstart{E}{ye} diseases present great challenges and serious threats to human health and quality of life. Vision impairments caused by many eye-related diseases such as glaucoma are irreversible and may result in complete vision loss if left untreated \cite{taylor2001world}. Detecting potential visual disorders at early stages is critical to reducing vision loss. In clinical practice, diagnosis of eye diseases largely relies on clinicians' assessments of ophthalmic images \cite{an2019glaucoma}, which have benefited from the development of various noninvasive medical imaging modalities such as the spectral domain optical coherence tomography (OCT). For example, the retinal nerve fiber layer (RNFL) thickness (RNFLT) map derived from OCT images is commonly used by clinicians to diagnose glaucoma and monitor disease progression \cite{wang2019machine}, based on the clinical knowledge that RNFL damage is a hallmark of glaucoma and therefore predictive of accompanying vision loss \cite{wang2020artificial}. 

Computer-aided diagnosis has been widely applied to ophthalmic images \cite{sengupta2020ophthalmic}, where the common paradigm is to develop effective feature learning algorithms that can automatically extract relevant biomarkers (e.g., optic nerve thinning and atrophy) to support the detection of eye disease and its progression. Recent works include 1) Using RNFLT maps to predict visual fields \cite{christopher2020deep,lazaridis2022predicting}; 2) Using fundus images to detect eye diseases \cite{das2021deep,saba2021automatic}; and 3) Using multiple image modalities such as fundus images and visual fields to collectively detect eye diseases and the progression \cite{an2019glaucoma,mursch2020artificial}. These studies demonstrate the promise of artificial intelligence (AI) algorithms in learning clinically meaningful features. Nevertheless, existing methods tend to learn sub-optimal features from ophthalmic images because of three limitations. First, current methods ignore the fact that ophthalmic images are commonly distorted by artifact due to segmentation failures in the context of
degraded imaging quality and differences in anatomy. For example, it has been reported that 35\% to 56\% of OCT images of the optic nerve has at least one artifact \cite{choi2020artifact}. Artifact may inhibit conventional AI methods from learning accurate features in ophthalmic images. Second, most current methods do not account for the fact that ophthalmic images are subject to variations in the testing procedure such as patient head rotation and imperfect fixation \cite{sengupta2020ophthalmic,mirzania2021applications}, which means these methods normally lack the mechanism to explicitly capture the intra- (i.e., feature invariance in the same image) and inter- (i.e., feature invariance between different images) image affinities. Third, current methods typically depend on massive quantities of image annotations to guide the model to learn features related to the target tasks, making these methods less generalizable across different tasks especially when the label information is not available.

\begin{figure}
  \centering
    \includegraphics[width=0.46\textwidth]{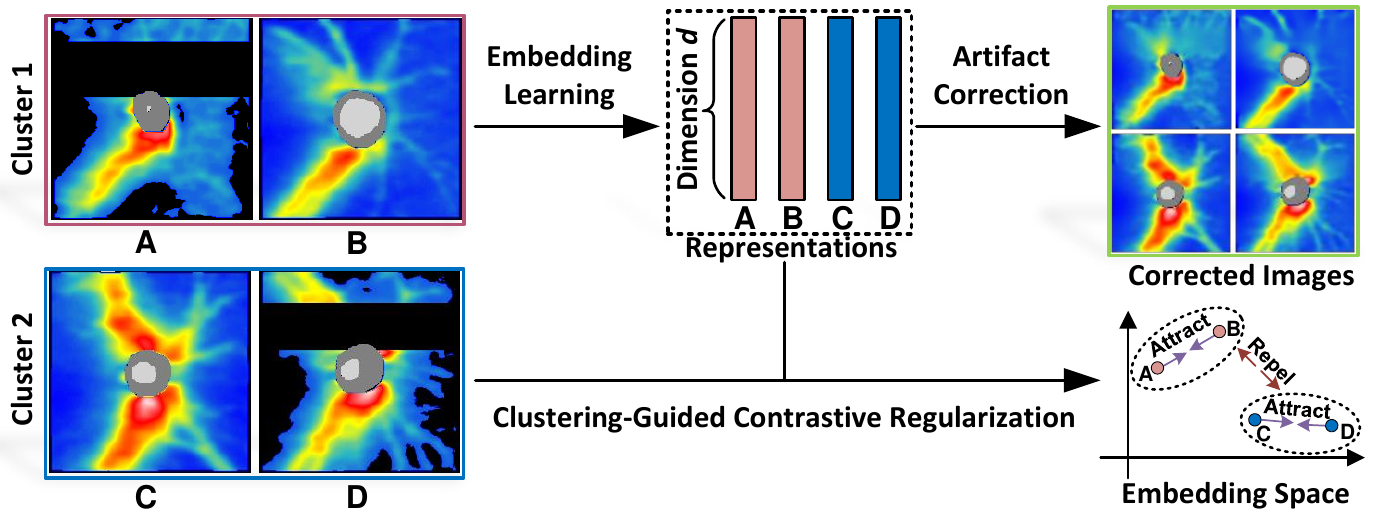}
  \caption{The embedding learning paradigm in \text{\modelname}. $A$, $B$, $C$ and $D$ are RNFLT maps in two clusters $\{A,B\}$ and $\{C,D\}$, where the optic disk rim and optic cup are colored in gray and light gray, respectively. The region in black color represents a segmentation artifact. \text{\modelname} learns the representations of ophthalmic images while reconstructing the pixels in areas affected by artifact. In addition, \text{\modelname} uses contrastive learning regularization to explicitly capture the relative affinities between images, where images from same (e.g., $A$ and $B$) or different clusters (e.g., $B$ and $C$) are encouraged to be represented with similar or dissimilar representations, respectively.} 
  \label{fig1}
\vspace{-3mm}
\end{figure}

Compared to prior work, we study a more generic question, namely feature representation or embedding learning of ophthalmic images. In this paper, representation learning and embedding refer to the same concept: learning a low-dimensional feature vector to represent each ophthalmic image with as much information preserved as possible, and also to explicitly embed the inter- and intra-image affinities. Embedding learning is a fundamental problem and can be relevant in multiple aspects. Intuitively, the well-learned representations of images are in a latent space that is machine-understandable and can be readily used as inputs to many lightweight machine learning algorithms to enhance their performance \cite{guo2016face}. In addition, the pretrained embedding learning model developed on one image type can be used as a starting point in developing other deep learning models set to be trained on other image types (i.e., transfer learning) \cite{wang2020dofe}. Other recognized benefits of embedding learning include facilitating the fusion of multi-modal data \cite{li2020self}, data interpretation and biomarker discovery \cite{gao2018interpretable}, and data visualization \cite{demiralp2014visual}. The key to embedding learning lies in extracting features from data such that both intra- and inter-data relationships can be accurately captured in a latent space. There are two challenges when applying embedding learning to ophthalmic images:
\begin{itemize}
    \item \textbf{Prevalent Image Artifacts:} Ophthalmic images are easily affected by artifacts, e.g., the RNFLT artifacts can result from the layer segmentation failure of OCT software. The image artifacts distort the derived features, resulting in the learned representations inaccurately representing the true image semantics.
    \item \textbf{Obscure Image Affinities:} Clinical interpretations of ophthalmic images are complicated primarily because of the subtle retinal anatomical variations between patients which can be difficult for clinicians to detect. The obscure inter-image affinities combined with unpredictable image artifacts make it challenging to learn distinguishing image representations.
\end{itemize}

We propose a general framework called \text{\modelname} for learning representations of ophthalmic images by taking the above two obstacles into account. As shown in Fig. \ref{fig1}, our strategy is to adopt an artifact-corrected embedding learning with contrastive embedding regularization to encourage distinguishing representations of ophthalmic images. Specifically, to address the first obstacle, we adopt an artifact correction-based embedding learning of the ophthalmic image, which forces the learned representation to reconstruct the complete image without artifact. To mitigate the second obstacle, we introduce a contrastive learning-based regularization, which encourages similar (i.e., within clusters) or dissimilar (i.e., between clusters) images to be represented with similar or dissimilar vectors. With this learning strategy of integrating artifact correction and contrastive regularization, \text{\modelname} is expected to output effective and discriminative representations of ophthalmic images that can be useful in relevant ophthalmic analytic tasks. We expect this to represent an improvement  over the state-of-the-art methods which fall into either supervised or unsupervised learning. 

In summary, the major contributions of this work are:
\begin{itemize}
    \item [1)] We proposed artifact-tolerant embedding learning for ophthalmic images. It is fundamental and suitable for understanding ophthalmic images which are often distorted by artifacts.
    \item [2)] We proposed \text{\modelname} which is a general framework for representation learning of ophthalmic images. It is tolerant to image artifact with an artifact correction module and it pursues discriminative representations of images with contrastive learning-based regularization.
    \item [3)] We evaluated \text{\modelname} with extensive experiments using a large OCT dataset of glaucoma patients through visual field prediction and glaucoma detection. \text{\modelname} is superior to state-of-the-art methods which adopt either supervised and unsupervised representation learning.
\end{itemize}

Section II surveys the related work. Section III presents preliminaries, including definition of the embedding learning problem and the building components used in our approach. The proposed model for embedding learning of ophthalmic images is introduced in Section IV. Section V reports on the results of experiments to validate the proposed approach. Finally, Section VI summarizes the conclusions we draw from this work.

\section{Related Work}
We first review feature representation learning for medical images. Then, we briefly present recent work on contrastive representation learning. Finally, we summarize related works on representation learning for ophthalmic images.

\subsection{Medical Image Representation Learning}
Representation learning in the medical domain is concerned with extracting useful information which help to guide diagnostic decision-making. Common medical image modalities include X-ray, magnetic resonance imaging (MRI), functional MRI, computed tomography, ultrasound scanning, optical coherence tomography, \textit{etc.} \cite{castiglioni2021ai}. There are two broad paths of research on this topic: self-supervised or unsupervised visual representation learning and supervised visual representation learning \cite{zhou2021review}. For self-supervised methods, pretext tasks such as image semantic reconstruction and pseudo label prediction are often used in order to achieve meaningful representations \cite{shurrab2021self}. For example, Bai \textit{et al.} \cite{bai2019self} proposed anatomical position prediction as a pretext task for cardiac MRI segmentation. Prakash \textit{et al.} \cite{prakash2020leveraging} adopted image denoising as a pretext task for nuclei images’ segmentation. Li \textit{et al.} \cite{li2020multi} combined two colorization based pretext tasks into a single multi-tasking framework called ColorMe to learn useful representations from scopy images. These pretext tasks can be roughly categorized into predictive, generative, contrastive and multi-tasking according to their working mechanisms \cite{shurrab2021self}. For example, generative methods normally learn to generate medical images using a generative adversarial network \cite{zhou2021review}, while multi-tasking methods seek to integrate various pretext tasks to co-optimize the representation learning \cite{li2020multi}. For supervised methods, representations are trained to explain the respective labels such as the binary class of benign and malignant tumors using the end-to-end task training. Popular tasks \cite{sarvamangala2021convolutional} include detection and severity classification of tumors, skin lesions, colon cancer, blood cancer, \textit{etc.}, achieved by various convolutional neural networks (CNN) architectures such as VGGNet \cite{simonyan2014very} and U-Net \cite{ronneberger2015u}.

Because the acquisition of quality annotations of medical images is expensive and depends heavily on clinicians' experience and knowledge, unsupervised representation learning has gained growing attention, especially in the current wave of research on generative and contrastive representation learning \cite{liu2021self}. Therefore, we adopt an unsupervised representation learning method in this work.

\subsection{Contrastive Representation Learning}
A recent breakthrough in contrastive learning has shed light on the potential of discriminative model for representation learning \cite{chen2020simple}. Contrastive learning aims at learning to compare diverse augmented images by optimizing the Noise-Contrastive Estimation loss function, where the positive pair is usually formed with two augmented views of the
same image, while negative ones are formed with different images \cite{shurrab2021self}. SimCLR \cite{chen2020simple} learns representations by maximizing agreement between differently augmented views of the same data example at the instance level. Contrastive clustering \cite{li2021contrastive} extends SimCLR to perform both instance and clustering level contrastive learning, where soft labels of instances are utilized to regularize affinities between samples. To address the reliance of most contrastive learning methods on a large number of explicit pairwise comparisons, SwAV \cite{caron2020unsupervised} simultaneously clusters the data while enforcing consistency between
cluster assignments produced for different augmentations. In a different way, InterCLR \cite{xie2020delving} uses a memory bank which stores running average features of all samples in the dataset computed in previous steps to reduce the limitation of small batch size. InterCLR additionally captures the inter-image invariance using \textit{k}-means clustering labels updated in mini-batch, where a positive sample for an input image is selected within the same cluster, while the negative samples are obtained from other clusters.

We adopt a similar learning paradigm as InterCLR to capture both the inter- and intra-invariance of ophthalmic images. However, we update clusters of images after every epoch which are thereafter used for guiding the contrastive learning in the following epoch. Since we use an artifact correction module as the backbone of \text{\modelname} to constrain the representation learning to be gradually optimized over the epochs, the clustering will become stable with the training to provide reliable guidance for contrastive learning.

\subsection{Ophthalmic Image Representation Learning}
Eye diseases such as glaucoma and diabetic retinopathy are chronic disorders that may result in irreversible blindness \cite{taylor2001world}. The past decade has seen the rise of AI methods to support diagnosis of eye diseases \cite{mirzania2021applications,sengupta2020ophthalmic}. For example, Wang \textit{al et.} \cite{wang2019machine} used  RNFLT maps (221 samples) to predict glaucoma with a CNN model. Nayak \textit{et al.} \cite{nayak2021ecnet} propose an evolutionary convolutional network model to predict glaucoma in fundus images (1,426 samples). An \textit{et al.} \cite{an2019glaucoma} used both the RNFLT maps and fundus images (357 samples) to predict glaucoma with a transfer learning-based CNN model. Since some ocular diseases are associated with functional vision loss, some work has attempted to predict the respective visual field damage using the ophthalmic image as input. For example, Lazaridis \textit{et al.} \cite{lazaridis2022predicting} used RNFLT maps (954 samples) to predict the visual fields with a multi-input CNN and a multi-channel variational autoencoder, respectively. Christopher \textit{et al.} \cite{christopher2020deep} used both RNFLT maps and optic nerve head en face images (1,909 samples) to predict visual field damage with a transfer learning-based ResNet model. However, these methods use relatively small sample sizes for model training and evaluation, which tend to generate over-fitting models and biased evaluation outcomes. In addition, they assume that the ophthalmic images are noise-free or simply exclude the noisy samples before training the models, which therefore cannot handle images containing artifacts.

The essence of AI-based diagnosis is to learn good feature representation revealing potential disease biomarkers in the ophthalmic image \cite{wang2020dofe,li2020self}. However, large image variations due to varying image artifact and patient-specific retinal anatomy make many standard deep learning methods suboptimal at learning quality representations. To mitigate this deficiency, we propose to correct the inclusive image artifact while learning its representation. We also introduce memory bank-supported contrastive learning to encourage distinguishable representations between images. Unlike many supervised methods which require a massive quantity of image annotations, our method is fully self-supervised which is more generalizable for extracting features for various types of ophthalmic images.

\begin{figure*}
  \centering
    \includegraphics[width=0.86\textwidth]{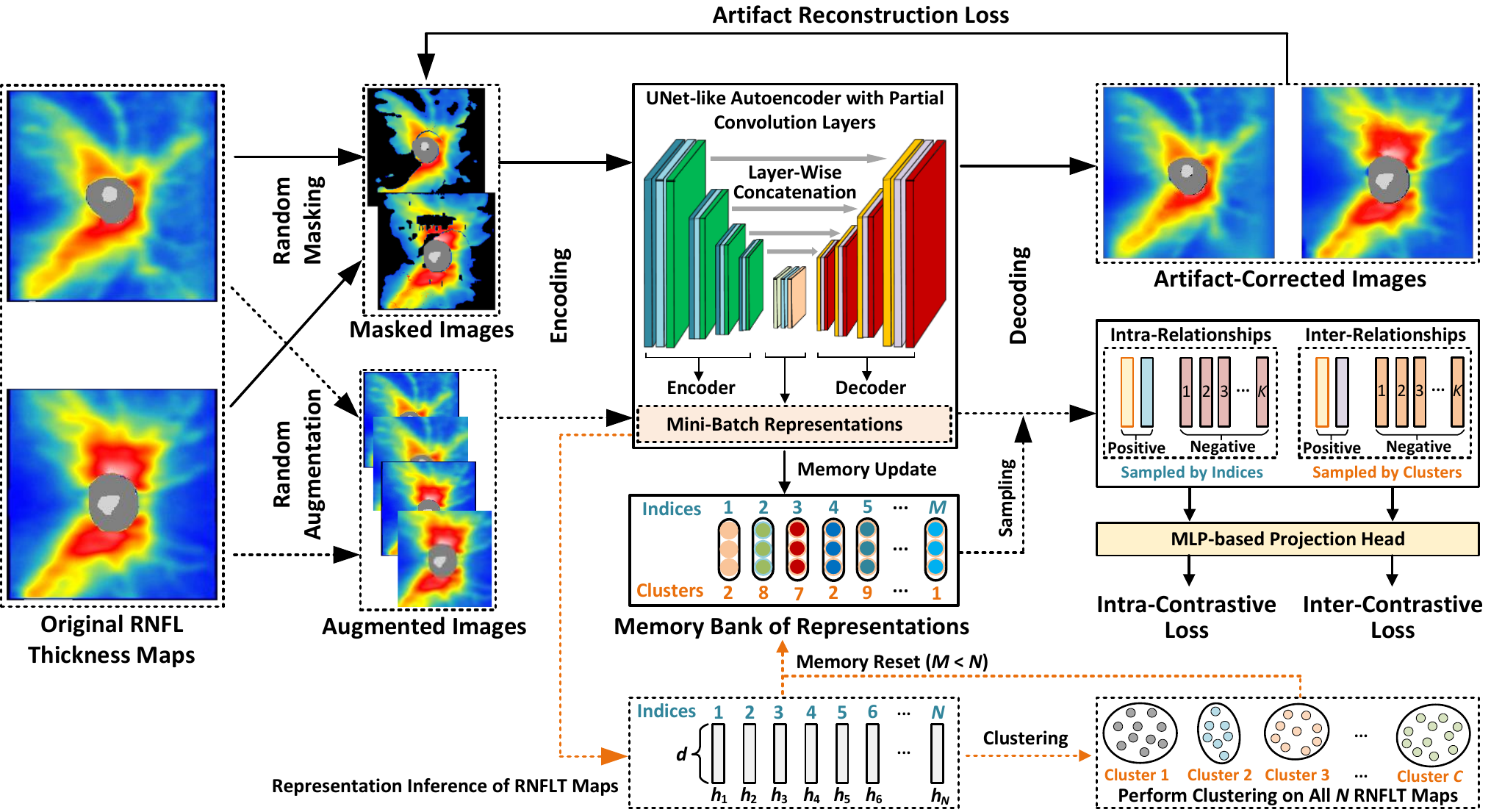}
  \caption{The proposed \text{\modelname} model for embedding learning of ophthalmic images (e.g., the RNFLT map).  \text{\modelname} contains an artifact correction-based embedding learning module and a contrastive learning-based regularization module to encourage discriminative embeddings. We apply a random mask on each input image to generate artifact pixels. The artifact correction module trains to correct the missing pixels at artifact locations in the masked image and meanwhile learns its vector representation by minimizing an artifact reconstruction loss. The contrastive learning-based regularization seeks to preserve the intra- and inter-image invariances in the embeddings by minimizing the intra-contrastive loss and the inter-contrastive loss, respectively. In addition, we use a memory bank to store the most recent feature representations of $\banksize$ samples. At the end of each epoch training, we perform clustering to obtain the cluster labels of images. These labels are updated to the memory bank to guide the selection of positive and negative pairs for intra- and inter-contrastive learning. The two learning modalities share the identical core which is an UNet-like autoencoder containing 16 partial convolution layers. Finally, the encoder outputs are taken as the representations of ophthalmic images.}
  \label{fig2}
\vspace{-2mm}
\end{figure*}

\section{Problem Definition and Preliminaries}
First, we formally define the embedding learning problem studied in this paper. Then, we briefly introduce the principle of partial convolution used as a building block in \text{\modelname} as well as the memory bank introduced to facilitate the contrastive learning.
\subsection{Problem Formulation}
Without loss of generality, we consider that each ophthalmic image is represented as a two-dimensional (2D) image, e.g., as a RNFLT map. Let $\data\in \spaceR^{\datasize\times \width\times \height}$ be a collection of $\datasize$ 2D ophthalmic images, where $\width$ and $\height$ are the width and height of the image, respectively. The embedding learning aims to project each image $\datax\in \data$ in a $\embeddim$-dimensional latent space $\embedspace\in \spaceR^{\embeddim}$, i.e., $\embed=f(\datax)$, where $f$ is the projection function defined as the embedding learning model \text{\modelname} (refer to Fig. \ref{fig2} for illustration) and $\embed\in \embedspace$ is the vector representation of $\datax$ with preserved clinical information conveyed by the image. In addition, we seek to improve embedding learning performance by preserving the relative affinities between ophthalmic images through the contrastive learning-based regularization in \text{\modelname}. 

\subsection{Partial Convolution}

Partial convolution \cite{liu2018image} is proposed to handle images which include missing pixels (e.g., holes). Let $\datax_{i,j}$ be the pixels within a convolution window at the location $(i,j)$ and $\maskm_{i,j}$ be the binary mask with valid pixels being all ones and invalid pixels being all zeros. The partial convolution at location $(i,j)$ is defined as:
\begin{equation}
  x^\prime_{i,j} =
    \begin{cases}
      \weightw^T(\datax_{i,j}\odot \maskm_{i,j})\scaler_{i,j} + \biasb, & \norm{\maskm_{i,j}}_1>0\\
      0, & \text{otherwise}
    \end{cases}       
\end{equation}
with 
\begin{equation}
    \scaler_{i,j}=\frac{\norm{\identity_{i,j}}_1}{\norm{\maskm_{i,j}}_1}
\end{equation}
where $\norm{\cdot}_1$ is the L1 norm, $\odot$ is the element-wise multiplication, $\weightw$ is the weight matrix and $\biasb$ is the bias. $\scaler_{i,j}$ is a scaling factor to adjust for the varying number of valid input pixels within the convolution window, where $\identity_{i,j}$ is a one-like matrix with the same shape as $\maskm_{i,j}$. We can observe from Eq. 1 that the output at every location depends only on the valid input pixels within the respective convolution window. This is achieved by maintaining a binary mask $\maskm_{i,j}$ for the feature map at each convolution layer. After a partial convolution layer $\layerl$, $\maskm_{i,j}$ at location ($i,j$) for next layer ($\layerl+1$) is updated as:
\begin{equation}
  m^\prime_{i,j} =
    \begin{cases}
      1, & \text{if } \norm{\maskm_{i,j}}_1>0\\
      0, & \text{otherwise}
    \end{cases}       
\end{equation}
meaning that if there exists at least one valid input pixel within the convolution window that contributes to the output $x^\prime_{i,j}$, the location ($i,j$) is a valid pixel for the next convolution layer; otherwise, it is an invalid pixel.

\subsection{Memory Bank}
During the contrastive learning process, the negative samples are normally constructed from samples within a  mini-batch. However, contrastive learning prefers a larger number of negative samples to learn quality representations \cite{xie2020delving,chen2020simple}. One approach would be to set a very large batch size, but this would require impractically large computational resources. As an alternative method, we use a memory bank of size $\banksize$ to store the representations and the clustering labels of most recent $\banksize$ running samples. Formally, a memory bank (i.e., a size-invariant queue) can be represented as $\membank=\{\bankfeat,\bankclust\}$, where $\bankfeat=\{\embed_i\}_{i=1:\banksize}$ is the up-to-date representations of $\banksize$ ophthalmic images and $\bankclust=\{\lab_i\}_{i=1:\banksize}$ is the clustering labels of images. In this paper, the image features are updated at the end of each training batch, while the clustering labels are updated at the end of each training epoch.

\section{Methodology}
The proposed framework for learning ophthalmic image representations, \text{\modelname}, is shown in Fig. 2. \text{\modelname} consists of an artifact correction-based representation learning module and a contrastive learning-based regularization module, where the latter encapsulates both an intra-contrastive learning and an inter-contrastive learning. Taken together, \text{\modelname} trains to optimize the following three parts:
\begin{itemize}
    \item \textbf{Artifact Correction-based Embedding Learning:} the backbone to reconstruct artifact pixels in masked areas of the input image and output its vector representation. This part trains to optimize an artifact reconstruction loss.  
    \item \textbf{Intra-Contrastive Learning-based Regularization:} applies image augmentations and encourages augmented images from the same image to learn similar representations while others to learn dissimilar representations. This part trains to optimize an intra-contrastive loss.
    \item \textbf{Inter-Contrastive Learning-based Regularization:} encourages images in the same clusters to learn similar representations, while images in different clusters to learn dissimilar representations. This part trains to optimize an inter-contrastive loss.
\end{itemize}

\subsection{Artifact Correction-based Embedding Learning}
Ophthalmic images are often affected by artifact or noise which prevents accurate interpretation and diagnosis of eye diseases. We propose to correct artifacts while learning the feature representation of images to address this problem. As shown in Fig. 2, the artifact correlation is implemented by an U-Net like structure \cite{ronneberger2015u} comprised of an encoder to learn the representation and a decoder to reconstruct the image from the learned representation. The encoder-decoder learning scheme can provide \text{\modelname} the potential to learn representation reflecting the corrected version of image without artifacts. 

Specifically, each encoder layer is a stack of three successive operation layers, including a partial convolution layer (refer to Section III-B for detail) to extract features from the masked image, a normalization layer to normalize the convolution feature map, and a non-linear activation layer (e.g., ReLU) to transform the feature map. In comparison, each decoder layer additionally includes an up-sampling layer and a concatenation layer before the partial convolution, normalization and non-linear activation layers. The concatenation operation concatenates two feature maps from the respective encoder and decoder layers at the same level. Notably, the last decoder layer's input contains the concatenation of the original masked image so the model can reuse valid pixels for image reconstruction. Formally, for each image $\datax$, the model takes the ground truth image $\datax_{gt}$, masked image $\datax_{in}$, and binary mask $\maskm$ (invalid pixels are zeros) as inputs. It then trains to minimize the differences with respect to both valid and invalid locations between the growth truth image $\datax_{gt}$ and the decoder-corrected image $\datax_{out}$ as:
\begin{equation}
    \loss_{valid} = \frac{1}{N_{\datax_{gt}}}\norm{\maskm\odot(\datax_{out}-\datax_{gt})}_1
\end{equation}

\begin{equation}
    \loss_{invalid} = \frac{1}{N_{\datax_{gt}}}\norm{(1-\maskm)\odot(\datax_{out}-\datax_{gt})}_1
\end{equation}
where $N_{\datax_{gt}}$ denotes the number of pixel locations in $\datax_{gt}$. In addition, several other losses have been proven useful in characterizing the similarity between the ground truth and predicted images \cite{liu2018image} in terms of different aspects of high-level features, which include the perceptual loss $\loss_{perceptual}$, the style loss $\loss_{style_{out}}$, the style loss $\loss_{style_{comp}}$ and the total variation loss $\loss_{tv}$. These losses are calculated from the outputs of an ImageNet-pretrained VGG-16 \cite{simonyan2014very} which takes $\datax_{gt}$ and $\datax_{out}$ as inputs. Formal definitions of these losses have been established in the literature \cite{liu2018image}. Taken together, the autoencoder training aims to optimize a collective reconstruction loss as:

\begin{equation}
\begin{aligned}
    \loss_{reconstruct} = & \loss_{valid} + \alpha\loss_{invalid} + \beta\loss_{perceptual}  \\
             &  + \gamma\loss_{style_{out}} + \delta\loss_{style_{comp}} + \eta\loss_{tv}
\end{aligned}
\end{equation}
Five weight parameters ($\alpha$, $\beta$, $\gamma$, $\delta$ and $\eta$) are used to balance the importance of different parts. Finally, the encoder output through an average pooling is taken as vector representation $\embed\in\spaceR^{\embeddim}$ for the input image $\datax_{in}$, where $\embeddim$ is the representation dimension.

\subsection{Intra-Contrastive Learning-based Regularization}
Intra-contrastive learning seeks to capture the intra-image invariance through data augmentations. Each image $\datax$ is augmented to generate two versions of variant images $\datax_i$ and $\datax_j$ which are then enforced to learn similar representations compared to other augmented images, i.e., they form a positive pair. In this paper, we adopt five different augmentation methods to randomly transform the image, including center crop, rotation, zooming, width shift and height shift. We design these augmentation types to account for the fact that ophthalmic image variances may be generated from varying conditions (e.g., distance of the eye to the lens and rotation of eyes) of patients when the images were taken. 

In addition, for each augmented image (e.g., $\datax_i$ or $\datax_j$), $\numnegs$ images are chosen to form negative pairs. Instead of constructing negative samples within each mini-batch as many previous works do \cite{chen2020simple,li2021contrastive}, we manage a memory bank (Fig. 2) implemented as a queue which stores the most recent representations of $\banksize$ ($\banksize>\numnegs$) images computed from the latest mini-batches. Then, we randomly sample $\numnegs$ images from the memory bank to form negative pairs for each augmented image. Following previous work \cite{chen2020simple}, we compute the normalized temperature-scaled cross entropy (NT-Xent) as the contrastive loss for each pair of positive examples ($i,j$) after a projection head:
\begin{equation}
    \loss_{intraCont} = -\log\frac{\exp{(\text{sim}(Pro(\embed_i),Pro(\embed_j))/\tau})}{\sum\limits^{\numnegs}_{n=1}\exp{(\text{sim}(Pro(\embed_i),Pro(\embed_n))/\tau})}
\end{equation}
where $\tau$ is a temperature parameter. $Pro(\cdot)$ is the projection head implemented as a three-layer multilayer perceptron (MLP).

\subsection{Inter-Contrastive Learning-based Regularization}
While intra-contrastive learning can capture the feature invariance between multiple augmented versions of an image, inter-contrastive learning seeks to explicitly model the relative affinities between different images. To this end, we adopt a clustering-guided image sampling strategy for constructing positive and negative samples. As shown in Fig. 2, we organize samples in clusters by performing the clustering (i.e., \textit{k}-means) after every epoch. Therefore, images will be assigned with respective clustering labels dynamically. Then, the newly updated labels after each epoch will be immediately reflected in the memory bank for the next epoch. For image $\datax$ in cluster $\lab_i$, we randomly sample an image $\datax'$ with label $\lab_i$ from the memory bank to form a positive pair with $\lab_i$. Then, we randomly sample $\numnegs$ images from other clusters to form negative pairs. Therefore, inter-contrastive learning tries to minimize the difference between images within the same clusters while encouraging different representation learning for images in different clusters. Finally, the inter-contrastive loss based on NT-Xent for each pair of positive ($u,v$) samples and $\numnegs$ pairs of negative samples $\{(u, n)\}_{n=1:\numnegs}$ is calculated as:
\begin{equation}
    \loss_{interCont} = -\log\frac{\exp{(\text{sim}(Pro(\embed_u),Pro(\embed_v))/\tau})}{\sum\limits^{\numnegs}_{n=1}\exp{(\text{sim}(Pro(\embed_u),Pro(\embed_n))/\tau})}
\end{equation}
Since we use the artifact correction module in \text{\modelname} as the backbone for learning representations of images, the clustering based on the learned representations will become stable with training, thus providing reliable guidance for contrastive sampling purpose. 

\subsection{Optimization of \text{\modelname}}
In \text{\modelname}, the artifact correction-based embedding learning and the contrastive learning-based regularization share the same learning core (i.e., the encoder), which collectively optimize the model to learn optimal representations of ophthalmic images under the combined objective as:  
\begin{equation}
    \loss= \sum\limits_{\datasize}(\loss_{reconstruct} + w_1\loss_{intraCont}+w_2\loss_{interCont})
\end{equation}
where $\datasize$ denotes the number of training samples, $w_1$ and $w_2$ are weight parameters used to balance the contributions of intra-contrastive and inter-contrastive learning, respectively. The entire training and optimizing procedure of \text{\modelname} is summarized in Algorithm 1.  

\SetInd{0.48em}{0.77em}
\begin{algorithm}[t]
\begin{small}
\DontPrintSemicolon
\SetAlgoLined
\SetKwInOut{Input}{Input}\SetKwInOut{Output}{Output}
\Input{Data $\data$ with $\datasize$ ophthalmic images, a set of masks $\masks$, clustering labels $\bankclust$ and memory bank $\membank$} 
\Output{The vector representation $\embed\in\spaceR^{\embeddim}$ for each $\datax\in\data$}
Initialize dimension $\embeddim$, weight parameters $w_1$ and $w_2$, training epochs $I$, batch size $\batchsize$, and number of clusters $\ncluster$.\\
\BlankLine
\For {$i \in [1,I]$}{
    \For{$j \in [1,\datasize // \batchsize]$}{
    $\data_b\leftarrow$ sample a batch of images from $\data$; \\
    $\maskm_b\leftarrow$ sample a batch of masks from $\masks$; \\
    $\data_m\leftarrow$ generate masked images from $\data_b$ and $\maskm_b$; \\
    $\data_a\leftarrow$ generate augmented images from $\data_b$; \\
    $\maskm_a\leftarrow$ generate masks covering the artifacts in $\data_a$; \\
    $\batchembeds\leftarrow$ generate representations from the encoder by \textbf{Encoder}($\data_m,\maskm_b,\data_a,\maskm_a$); \\
    $\membank\leftarrow$ update the memory bank with $\batchembeds$;\\
    $\data_o\leftarrow$ generate corrected images from the decoder by \textbf{Decoder}($\data_m,\maskm_b,\data_a,\maskm_a$); \\
    $\loss_{reonstruct}\leftarrow$ compute the reconstruction loss based on $\data_b,\maskm_b,\data_o$ by Eq. 6; \\
    $\data_{intra}\leftarrow$ sample negative images from $\membank$;\\
    $\loss_{intraCont}\leftarrow$ compute the intra-contrastive learning loss based on $\data_a,\maskm_a,\batchembeds$ and $\data_{intra}$ by Eq. 7; \\
    $\data_{inter}\leftarrow$ sample positive and negative images from $\membank$;\\
    $\loss_{interCont}\leftarrow$ compute the inter-contrastive learning loss based on $\data_m,\maskm_b,\batchembeds$ and $\data_{inter}$ by Eq. 8; \\
    Minimize loss $\loss$ in Eq. 9 by the Adam optimizer.
    }
    $\embedspace\leftarrow$ output the representations from \textbf{Encoder}($\data,\masks$)
}
return $\embedspace$
\caption{Training the \text{\modelname} model}
\end{small}
\end{algorithm}

\begin{table*}[ht]
\renewcommand{\arraystretch}{1.28}
\centering
\caption{The visual field prediction performance based on the ridge regression model (bold-faced results are statistically significantly different from baseline methods RPCA, DAE and SimCLR by a paired \textit{t}-student test with $p \leq 0.05$).}
\begin{adjustbox}{width=0.96\textwidth}
\small
\begin{tabular}{l|cc|cc|cc|cc|cc}
\toprule
\multicolumn{1}{c|}{Training Ratios} & \multicolumn{2}{c|}{10\%} & \multicolumn{2}{c|}{30\%} &\multicolumn{2}{c|}{50\%} & \multicolumn{2}{c|}{70\%} &
\multicolumn{2}{c}{90\%}\\
\hline
\multicolumn{1}{c|}{Methods} & MAE & R$^2$ & MAE & R$^2$ & MAE & R$^2$ & MAE & R$^2$ & MAE & R$^2$\\
\hline
RPCA & {4.373{\tiny $\pm0.18$}} & {0.111{\tiny $\pm0.07$}} & {3.669{\tiny $\pm0.05$}} & {0.341{\tiny $\pm0.013$}}& {3.648{\tiny $\pm0.03$}} & {0.382{\tiny $\pm0.01$}} & {3.580{\tiny $\pm0.02$}} &  {0.403{\tiny $\pm0.01$}} & {3.344{\tiny $\pm0.01$}} & {0.456{\tiny $\pm0.00$}} \\

DAE & {3.426{\tiny $\pm0.09$}} & {0.439{\tiny $\pm0.03$}} & {3.293{\tiny $\pm0.06$}} & {0.476{\tiny $\pm0.02$}} & {3.199{\tiny $\pm0.05$}} &  {0.501{\tiny $\pm0.02$}} & {3.161{\tiny $\pm0.05$}} & {0.505{\tiny $\pm0.02$}} & {3.119{\tiny $\pm0.04$}} & {0.523{\tiny $\pm0.02$}}\\

SimCLR & {3.494{\tiny $\pm0.08$}} & {0.392{\tiny $\pm0.03$}} & {3.399{\tiny $\pm0.05$}} & {0.427{\tiny $\pm0.02$}} & {3.350{\tiny $\pm0.05$}} &  {0.437{\tiny $\pm0.02$}} & {3.344{\tiny $\pm0.05$}} & {0.443{\tiny $\pm0.02$}} & {3.337{\tiny $\pm0.06$}} & {0.448{\tiny $\pm0.02$}}\\
\hline

\text{\modelname$_{recon}$} & {3.262{\tiny $\pm0.06$}} & {0.460{\tiny $\pm0.02$}} & {3.142{\tiny $\pm0.03$}} & {0.495{\tiny $\pm0.01$}} & {3.100{\tiny $\pm0.02$}} &  {0.506{\tiny $\pm0.01$}} & {3.078{\tiny $\pm0.02$}} & {0.513{\tiny $\pm0.01$}} & {3.065{\tiny $\pm0.01$}} & {0.517{\tiny $\pm0.01$}}\\

\text{\modelname$_{intra}$} & {3.452{\tiny $\pm0.06$}} & {0.424{\tiny $\pm0.01$}} & {3.329{\tiny $\pm0.03$}} & {0.451{\tiny $\pm0.01$}} & {3.277{\tiny $\pm0.02$}} &  {0.461{\tiny $\pm0.01$}} & {3.254{\tiny $\pm0.01$}} & {0.466{\tiny $\pm0.01$}} & {3.235{\tiny $\pm0.01$}} & {0.470{\tiny $\pm0.01$}}\\

\text{\modelname$_{inter}$} & {3.201{\tiny $\pm0.06$}} & {0.478{\tiny $\pm0.01$}} & {3.079{\tiny $\pm0.03$}} & {0.508{\tiny $\pm0.01$}} & {3.038{\tiny $\pm0.02$}} &  {0.518{\tiny $\pm0.01$}} & {3.019{\tiny $\pm0.01$}} & {0.522{\tiny $\pm0.01$}} & {3.002{\tiny $\pm0.01$}} & {0.526{\tiny $\pm0.01$}}\\

\text{\modelname} & {\textbf{3.079}{\tiny $\pm0.07$}} & {\textbf{0.494}{\tiny $\pm0.02$}} & {\textbf{3.002}{\tiny $\pm0.04$}} & {\textbf{0.519}{\tiny $\pm0.01$}} & {\textbf{2.963}{\tiny $\pm0.03$}} &  {\textbf{0.529}{\tiny $\pm0.01$}} & {\textbf{2.948}{\tiny $\pm0.02$}} & {\textbf{0.535}{\tiny $\pm0.01$}} & {\textbf{2.933}{\tiny $\pm0.01$}} & {\textbf{0.540}{\tiny $\pm0.01$}}\\
\bottomrule
\end{tabular}
\end{adjustbox}
\vspace{-2mm}
\end{table*}

\begin{table*}[ht]
\renewcommand{\arraystretch}{1.28}
\centering
\caption{The glaucoma detection performance (\%) based on the linear SVC (bold-faced results are statistically significantly different from baseline methods RPCA, DAE and SimCLR by a paired \textit{t}-student test with $p \leq 0.05$).}
\begin{adjustbox}{width=0.96\textwidth}
\small
\begin{tabular}{l|cc|cc|cc|cc|cc}
\toprule
\multicolumn{1}{c|}{ Ratios} & \multicolumn{2}{c|}{10\%} & \multicolumn{2}{c|}{30\%} &\multicolumn{2}{c|}{50\%} & \multicolumn{2}{c|}{70\%} &
\multicolumn{2}{c}{90\%}\\
\hline
\multicolumn{1}{c|}{Methods} & Accuracy & F1 & Accuracy & F1 & Accuracy & F1 & Accuracy & F1 & Accuracy & F1\\
\hline
RPCA & {68.05{\tiny $\pm1.94$}} & {68.94{\tiny $\pm2.09$}} & {77.24{\tiny $\pm0.77$}} & {78.51{\tiny $\pm0.86$}} & {77.03{\tiny $\pm0.63$}} &  {78.12{\tiny $\pm0.84$}} & {78.88{\tiny $\pm0.51$}} & {79.71{\tiny $\pm0.67$}} & {79.67{\tiny $\pm0.32$}} & {80.92{\tiny $\pm0.34$}}\\

DAE & {76.70{\tiny $\pm1.51$}} & {77.63{\tiny $\pm1.62$}} & {77.63{\tiny $\pm0.80$}} & {79.37{\tiny $\pm0.96$}} & {79.34{\tiny $\pm0.82$}} &  {80.16{\tiny $\pm0.90$}} & {80.26{\tiny $\pm0.82$}} & {80.72{\tiny $\pm0.96$}} & {80.50{\tiny $\pm0.76$}} & {81.36{\tiny $\pm0.72$}}\\

SimCLR & {75.57{\tiny $\pm1.09$}} & {76.83{\tiny $\pm1.39$}} & {77.27{\tiny $\pm0.79$}} & {78.92{\tiny $\pm0.84$}} & {77.71{\tiny $\pm0.80$}} &  {79.38{\tiny $\pm0.83$}} & {77.88{\tiny $\pm0.71$}} & {79.39{\tiny $\pm0.75$}} & {77.98{\tiny $\pm0.78$}} & {79.84{\tiny $\pm0.80$}}\\
\hline

\text{\modelname$_{recon}$} & {77.56{\tiny $\pm0.82$}} & {78.96{\tiny $\pm1.16$}} & {79.16{\tiny $\pm0.50$}} & {80.70{\tiny $\pm0.57$}} & {79.65{\tiny $\pm0.39$}} &  {81.25{\tiny $\pm0.37$}} & {79.99{\tiny $\pm0.39$}} & {81.55{\tiny $\pm0.36$}} & {80.20{\tiny $\pm0.26$}} & {81.76{\tiny $\pm0.23$}}\\

\text{\modelname$_{intra}$} & {78.04{\tiny $\pm0.66$}} & {80.72{\tiny $\pm0.30$}} & {79.53{\tiny $\pm0.34$}} & {81.69{\tiny $\pm0.24$}} & {80.08{\tiny $\pm0.28$}} &  {82.12{\tiny $\pm0.24$}} & {80.36{\tiny $\pm0.20$}} & {82.30{\tiny $\pm0.19$}} & {80.53{\tiny $\pm0.13$}} & {82.44{\tiny $\pm0.11$}}\\

\text{\modelname$_{inter}$} & {79.18{\tiny $\pm0.67$}} & {80.67{\tiny $\pm0.87$}} & {80.50{\tiny $\pm0.55$}} & {82.03{\tiny $\pm0.58$}} & {81.10{\tiny $\pm0.34$}} &  {82.61{\tiny $\pm0.37$}} & {81.34{\tiny $\pm0.33$}} & {82.80{\tiny $\pm0.34$}} & {81.50{\tiny $\pm0.21$}} & {82.95{\tiny $\pm0.22$}}\\

\text{\modelname} & {\textbf{79.91}{\tiny $\pm0.64$}} & {\textbf{81.57}{\tiny $\pm0.75$}} & {\textbf{80.82}{\tiny $\pm0.41$}} & {\textbf{82.37}{\tiny $\pm0.45$}} & {\textbf{81.19}{\tiny $\pm0.31$}} &  {\textbf{82.73}{\tiny $\pm0.36$}} & {\textbf{81.47}{\tiny $\pm0.22$}} & {\textbf{82.97}{\tiny $\pm0.24$}} & {\textbf{81.57}{\tiny $\pm0.15$}} & {\textbf{83.09}{\tiny $\pm0.15$}}\\

\bottomrule
\end{tabular}
\end{adjustbox}
\vspace{-3mm}
\end{table*}

\section{EVALUATION METHODS AND RESULTS}
We evaluated \text{\modelname} using a real-world ophthalmic image dataset, i.e., RNFLT maps of glaucoma patients. We compare \text{\modelname} with prior work for glaucoma detection and visual field prediction. Based on the learned representations, visual field prediction aims to predict mean defect level of vision loss, while glaucoma detection aims to predict the binary glaucomatous or non-glaucomatous status of patients.

\subsection{Datasets}
The dataset includes 30,953 RNFLT maps (each containing $225\times 225$ pixels) from 11,284 unique patients who received a clinical diagnosis between 2010 and 2021. There are numbers of 14,730 left-eye RNFLT maps and 16,223 right-eye RNFLT maps. All left-eye RNFLT maps were horizontally flipped to the right-eye format. Pretraining of the \text{\modelname} model was done using 22,953 RNFLT maps; another 2,000 RNFLT maps were used for parameter selection. The remaining 6,000 RNFLT maps were used for evaluating the glaucoma detection and visual field mean deviation (MD) prediction.

\subsection{Comparative Methods}
Glaucoma detection and visual field prediction with deep learning methods have received increasing attention \cite{li2019attention,mirzania2021applications}, but few of existing methods focus on the unsupervised feature representation learning. We choose to compare the following high-performing methods that are relevant to our method for unsupervised representation learning of RNFLT maps:
\begin{itemize}
    \item \textbf{RPCA} \cite{candes2011robust}. Robust principal component analysis (PCA) is a modified version of the widely used PCA method which works well with respect to noisy or corrupted data.
    \item \textbf{DAE} \cite{vincent2010stacked}. Denoising autoencoder (AE) extends from the AE by changing the reconstruction criterion. It is effective at learning representations of noisy or corrupted data.
    \item \textbf{SimCLR} \cite{chen2020simple}. This is a contrastive learning method which performs data augmentations and forces similar or dissimilar images in order to learn similar or dissimilar representations.
    \item \textbf{\modelname}. This is our proposed artifact-corrected representation learning combined with intra-contrastive and inter-contrastive learning-based regularization.
\end{itemize}

We perform glaucoma detection and visual field detection based on the learned representations by the above methods.
We further compare \text{\modelname} with several state-of-the-art visual representation learning methods in two categories which are designed for end-to-end supervised classification training, including the vision transformer which has been used for relevant glaucoma detection tasks \cite{song2021deep}:
\begin{itemize}
    \item \textbf{Vision Transformer (ViT)}. This is a type of deep neural network mainly based on the self-attention mechanism which has achieved the state of the art in many image learning tasks. Here, we include several recent strong transformers, CCT \cite{hassani2021escaping}, ViT \cite{dosovitskiy2020image} and CrossVit \cite{chen2021crossvit}, for comparison.
    \item \textbf{Vision Graph Neural Networks (GNN)}. This is an recent vision feature learning model which models an image as a graph of small patches and uses GNN to capture the semantic relevance between neighborhood batches \cite{han2022vision}.
\end{itemize}

Because vision transformer and vision GNN are used for supervised image classification, we compare their performance through the binary glaucoma detection task in the experiment.

\subsection{Experimental Parameters}
For glaucoma detection, we train a linear support vector classifier (SVC) based on learned representations and adopt the Accuracy (Acc) and F1 score as evaluation metrics. For visual field (i.e., MD) prediction, we train a ridge regression model and adopt the mean absolute error (MAE) and R-squared score (R$^2$) as evaluation metrics. We repeat the evaluations 40 times with random partitions of various ratios of training data. The mean performances and standard deviations are reported. 

We train the \text{\modelname} model 80 epochs with a batch size of 4 and adopt the following parameter settings. The learning rate for the Adam optimizer is 0.0002.  In Eq. 6, $\alpha$, $\beta$, $\gamma$, $\delta$ and $\eta$ are set as 6, 0.05, 1, 1 and 0.1, respectively. In Eq. 9, $w_1$ and $w_2$ are set as 0.002 and 0.001, respectively. The representation dimension $\embeddim$ is set as 512. The number of image clusters is set as 7. The bank size $\banksize$ and number of negative samples $\numnegs$ are set as 800 and 16, respectively. Detailed parameter settings are referred to our implementation of the \text{\modelname} model\footnote{https://github.com/codesharea/EyeLearn.}.

\begin{table}[ht]
\renewcommand{\arraystretch}{1.28}
\centering
\caption{The comparison with supervised vision transformer and vision GNN on glaucoma detection performance (\%).}
\begin{adjustbox}{width=0.5\textwidth}
\small
\begin{tabular}{c|c|ccc}
\toprule
\multicolumn{1}{c|}{Categories} & Methods & Accuracy & F1 & FLOPs \\
\hline
\multirow{3}{*}{Vision Transformer} & CCT \cite{hassani2021escaping} & {80.39} & {81.39} & {1.30B} \\
 & ViT \cite{dosovitskiy2020image} & {78.78} & {80.86} & {0.14B} \\
 & CrossViT \cite{chen2021crossvit} & {76.06} & {79.31} & {0.13B} \\
\hline

\multirow{1}{*}{Vision GNN} & ViG \cite{han2022vision} & {81.39} & {82.88} & {4.56B} \\
\hline

\multirow{2}{*}{Our Method} & EyeLearn (with SVC) & {81.44} & {82.78} & {0.16B} \\
 & EyeLearn (with 2-Layer MLP) & {81.50} & {82.67} & {0.16B} \\

\bottomrule
\end{tabular}
\end{adjustbox}
\end{table}

\subsection{Results}
In this section, we first report the embedding performance of various baseline methods by the downstream visual field prediction and glaucoma detection tasks. Then, we examine the contributions of various learning components in \text{\modelname} through the ablation study. Finally, we compare the performance at various proportions of artifacts in the RNFLT maps.
\subsubsection{Visual Field Prediction}
The visual field prediction results are shown in Table 1. We can summarize three major observations: a) Deep learning models such as DAE, SimCLR and \text{\modelname} perform better than the conventional model RPCA to learn representations from RNFLT maps. This is attributed to the  efficient feature learning and optimization strategies adopted in deep learning models; b) The autoencoder-based methods, i.e., \text{\modelname} and DAE, generally outpace the purely contrastive learning model SimCLR for learning representations of RNFLT maps. This is possibly because RNFLT maps share many similar features, making contrastive learning alone ineffective to discriminate between images. In contrast, the autoencoder trains to reconstruct the input features from the learned representation which is robust to achieve meaningful representations; c) Our method has better generalizability (i.e., stable performance) when the training ratios changes from 10\% to 90\%. For example, \text{\modelname} can achieve better prediction performance with fewer data (i.e., 10\%) compared with other methods trained on more data (i.e., 30\%). Our method integrating both the autoencoder and contrastive learning performs the best. The average MAE and R$^2$ of \text{\modelname} respectively improved by 0.255 and 3.4\% over DAE, and 0.400 and 9.4\% over SimCLR, which verifies the superiority of \text{\modelname} for the embedding learning of ophthalmic images. 

\begin{figure}
  \centering
    \includegraphics[width=0.48\textwidth]{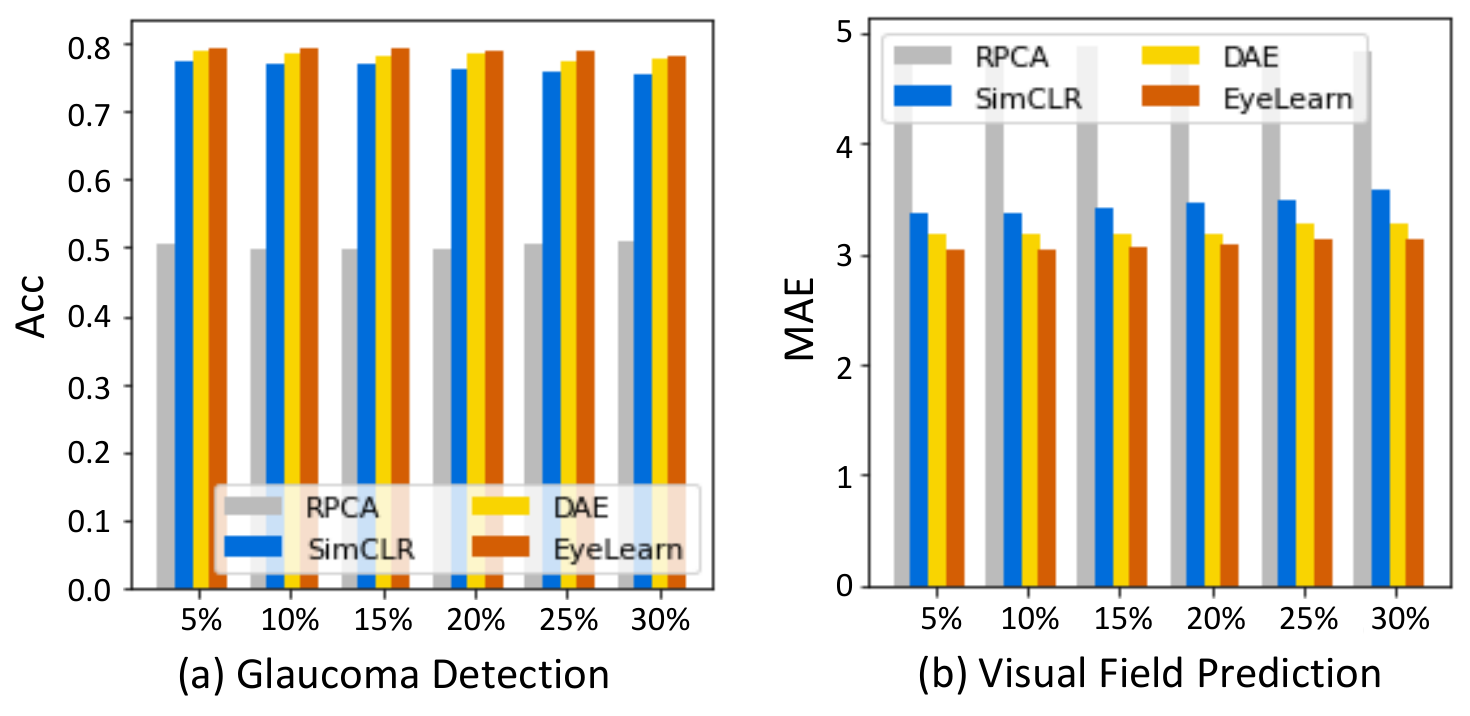}
  \caption{Performance at various proportions of RNFLT artifacts.}
  \label{fig6}
\vspace{-3mm}
\end{figure}

\subsubsection{Glaucoma Detection}
The glaucoma detection results for various unsupervised methods are shown in Table 2. We can observe that DAE and \text{\modelname} outperform other baseline methods, which again verifies the advantage of autoencoder-based embedding learning of ophthalmic images. For glaucoma detection, it is interesting to observe that SimCLR is not better than RPCA when the training ratio is 90\% as in the visual field prediction, meaning the learned representations by SimCLR do not generalize well in different downstream tasks. Among the algorithms, \text{\modelname} has achieved the best prediction performance, which demonstrates its superiority and the learned representations are well-generalized. In addition, \text{\modelname} has achieved better detection performance using less training data (i.e., 10\%) compared to RPCA, DAE and SimCLR which require more data (i.e., 90\%) to achieve comparative results. This means \text{\modelname} has better robustness and generalizability than other methods.

We further compare with state-of-the-art supervised embedding learning methods and the results are shown in Table 3. We train a SVC classifier and a 2-layer MLP classifier for glaucoma detection based on the previously learned representations by \text{\modelname} in an unsupervised manner. We can observe that our methods are superior to the vision transformers and can achieve comparable performance (Accuracy and F1 score) to the vision GNN. Floating point operations per second (FLOPs) is a vital measure of the model computational efficiency. We can observe from Table 3 that \text{\modelname} requires fewer FLOPs compared to ViG and CCT which are the top-two supervised embedding learning methods.

\begin{figure}
  \centering
    \includegraphics[width=0.49\textwidth]{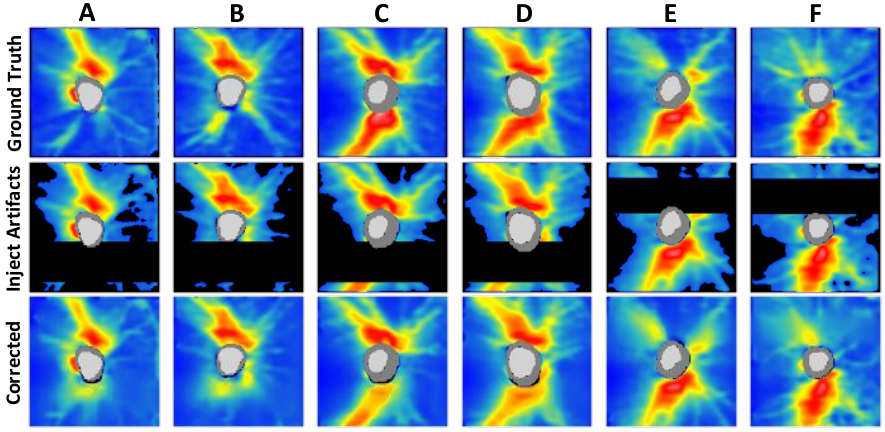}
  \caption{Examples of the artifact correction effect demonstrating the similarity between \text{\modelname} corrected maps and the ground truth. The first row is the original RNFLT maps. The second row is the respective maps with injected artifacts. The last row is the corrected maps by \text{\modelname}.} 
  \label{fig7}
\vspace{-3mm}
\end{figure}

\begin{figure}
  \centering
    \includegraphics[width=0.4\textwidth]{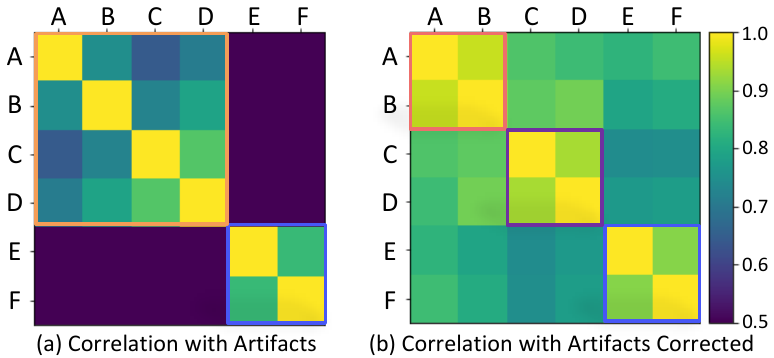}
  \caption{The pairwise correlations between RNFLT maps in the (a) second row and (b) third row shown in Fig. \ref{fig7}.}
  \label{fig8}
\vspace{-3mm}
\end{figure}

\begin{figure*}
  \centering
    \includegraphics[width=1\textwidth]{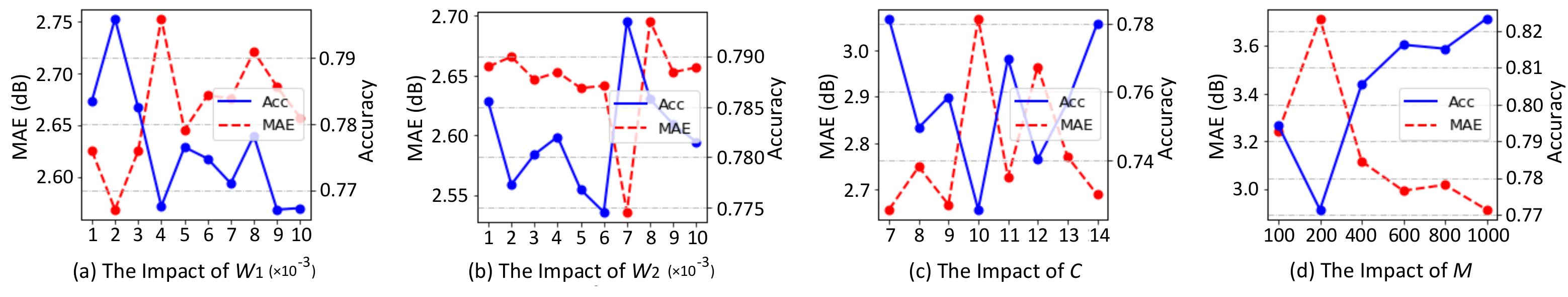}
  \caption{The impacts of the weight parameters $w_1$ and $w_2$, the number of clusters $\bankclust$ and the bank size $\banksize$.}
  \label{fig5}
\vspace{-3mm}
\end{figure*}

\subsubsection{Ablation Study}
We develop three variants of \text{\modelname} to examine the benefits of its different learning parts. \text{\modelname$_{recon}$} is the version created by removing the contrastive learning-based regularization. \text{\modelname$_{intra}$} is the version created by removing the inter-contrastive learning-based regularization. \text{\modelname$_{inter}$} is the version created by removing the intra-contrastive learning-based regularization. The performance of these variant methods with respect to visual field prediction and glaucoma detection is shown in Table 1 and Table 2. We can observe that \text{\modelname$_{intra}$} performs better than \text{\modelname$_{recon}$} in glaucoma detection but not in visual field prediction. This is possibly because intra-contrastive learning aims to capture the intra-feature invariance in an image, ignoring the inter-image feature invariance which is important for learning representations used in the visual field prediction task. In contrast, \text{\modelname$_{inter}$} outperforms \text{\modelname$_{recon}$} which demonstrates that capturing inter-image invariance is useful for learned improved embeddings. However, simultaneously capturing the intra-feature invariance and inter-image invariance delivers the best embedding learning model (\text{\modelname}) as observed in the comparisons in Tables 1 and 2.

\subsubsection{Comparison at Different Proportions of Artifact}
We generate different proportions of artifact noise into the RNFLT maps before inferring their representations with the pretrained models. We can observe from the comparative results in Fig. 3 that \text{\modelname} consistently performs better than other methods and demonstrates relatively stable performance for different percentages of artifacts. The reason is that \text{\modelname} is designed to handle image artifacts so that the learned representation is likely to reflect the image with artifacts corrected.

\subsection{Case Study}
We examine the impact of correcting image artifact with the example images in three categories $\{A,B\}$, $\{C,D\}$ and $\{E,F\}$ (Fig. 4). We can observe that \text{\modelname} can impute visually meaningful pixels at the artifact regions. For example, the lower thickness bundles in A and B are missing, which are roughly predicted by our model. We further obtain the pairwise Pearson correlations between images in the second row (i.e., calculated based on the raw features in the image) and third rows (i.e., calculated based on the learned representations by \text{\modelname}). The results are shown in Fig. 5. We can observe that images with artifacts fall in two categories $\{A,B,C,D\}$ and $\{E,F\}$, while \text{\modelname} tends to recover the three true categories with the artifacts corrected. In addition, after correcting the image artifacts, the pairwise correlations of $\{A,B\}$, $\{C,D\}$ and $\{E,F\}$ become closer, as can been seen by comparing the color maps between Fig. 5 (a) and (b). This means that \text{\modelname} is able to learn closed representations for similar images and recover their true relationships that have been distorted by artifacts.

\subsection{Parameter Sensitivity}
We study the impacts of important hyperparameters including the weight parameters $w_1$ and $w_2$ used in Eq. 9, the number of clusters $\ncluster$ in the clustering-guided contrastive learning, and the size of the memory bank $\banksize$. We can observe from Fig. \ref{fig5} that the variances of these parameters have significant impacts to the model performance. For example, \text{\modelname} tends to achieve improved embedding performance with a larger memory bank size observed from the trend in Fig. \ref{fig5} (d).

\section{Conclusion}
Learning the feature representations of ophthalmic images is fundamental to computer-aided eye disease diagnosis. Recent work has explored the potential to extract biomarker features from ophthalmic images using AI techniques for detecting and monitoring the progression of glaucoma. The challenges for embedding ophthalmic images include prevalent image artifacts and obscure image affinities. To address these problems, we propose \text{\modelname}, which is a self-supervised framework for representation learning of ophthalmic images. \text{\modelname} adopts an artifact correction-based representation learning, which enables the learned representation to reflect the corrected image without artifacts. Another feature in \text{\modelname} is the contrastive learning-based regularization used to encourage improved and discriminative representations between ophthalmic images. We follow a clustering-guided contrastive learning which emphasizes the intra-feature invariance through image augmentation and the inter-image feature invariance by organizing images in clusters (i.e., images in the same clusters learn similar representations). 

We designed comprehensive experiments to evaluate the learned feature representations by \text{\modelname} using a large ophthalmic image dataset of glaucoma patients. Experimental results show that \text{\modelname} outperforms strong unsupervised representation learning models by a large margin. The comparison also shows that \text{\modelname} performs comparably or better than some state-of-the-art supervised methods regarding the model effectiveness and computational efficiency. The case study further demonstrated that \text{\modelname} is useful for correcting artifacts and learn the actual relations between ophthalmic images.


\bibliographystyle{IEEEtran}
\bibliography{references}

\end{document}